\documentstyle[a4]{article}
  \newlength\titlebox 
 \twocolumn 
\def\addcontentsline#1#2#3{}

\skip\footins 9pt plus 4pt minus 2pt
\def\footnoterule{\kern-3pt \hrule width 5pc \kern 2.6pt }
\labelwidth\leftmargini\advance\labelwidth-\labelsep \labelsep 5pt

\def\@listi{\leftmargin\leftmargini}
\def\@listii{\leftmargin\leftmarginii
   \labelwidth\leftmarginii\advance\labelwidth-\labelsep
   \topsep 2pt plus 1pt minus 0.5pt
   \parsep 1pt plus 0.5pt minus 0.5pt
   \itemsep \parsep}
\def\@listiii{\leftmargin\leftmarginiii
    \labelwidth\leftmarginiii\advance\labelwidth-\labelsep
    \topsep 1pt plus 0.5pt minus 0.5pt
    \parsep \z@ \partopsep 0.5pt plus 0pt minus 0.5pt
    \itemsep \topsep}
\def\@listiv{\leftmargin\leftmarginiv
     \labelwidth\leftmarginiv\advance\labelwidth-\labelsep}
\def\@listv{\leftmargin\leftmarginv
     \labelwidth\leftmarginv\advance\labelwidth-\labelsep}
\def\@listvi{\leftmargin\leftmarginvi
     \labelwidth\leftmarginvi\advance\labelwidth-\labelsep}

\belowdisplayskip \abovedisplayskip
\title{A State-Transition Grammar for Data-Oriented
   Parsing}
\author{David Tugwell\thanks{This research was funded by a research
    studentship from the ESRC. My thanks also for discussion and
comments to Matt Crocker, Chris Brew, David Milward and Anna Babarczy.}\\
 Centre for Cognitive Science,
 University of Edinburgh\\
 2, Buccleuch Place,
 Edinburgh EH8 9LW, Scotland\\
 davidt@cogsci.ed.ac.uk\\}

\date{}
\begin{document}

\maketitle

\begin{abstract}
\vspace{-5pt}
This paper presents a grammar formalism designed for use in data-oriented
approaches to language processing. It goes on to investigate
ways in which a corpus pre-parsed with this
formalism may be processed to provide a probabilistic language model
for use in the parsing of fresh texts.
\end{abstract}
\vspace{-3pt}

\section{Introduction}
\vspace{-5pt}

Recent years have seen a resurgence of interest in
probabilistic techniques for automatic language analysis. In
particular, there has arisen a distinct paradigm of processing on the basis of
pre-analyzed data which has taken the name {\bf
  Data-Oriented Parsing}.

\vspace{-3pt}
\begin{quotation}
``Data Oriented Parsing (DOP) is a model where no abstract rules, but
language experiences in the form of an analyzed corpus, constitute the
basis for language processing.''\footnote{Bod, 1992.}
\end{quotation}
\vspace{-3pt}

There is not space here to present full justification for
adopting such an approach or to detail the advantages that it
offers. The main claim it makes is that effective language processing requires
a consideration of both the structural and statistical aspects of
language, whereas traditional competence grammars rely only on the
former, and standard  statistical techniques such as n-gram models
only on the latter.  DOP attempts to combine
these two traditions and produce ``performance grammars'', which:

\vspace{-3pt}
\begin{quotation}
``... should not only contain information on
the structural possibilities of the general language system, but also
on details of actual language use in a language
community...''\footnote{ibid.}
\end{quotation}
\vspace{-3pt}

This approach entails however that a corpus has first to be
pre-analyzed (ie.\ hand-parsed), and the question immediately arises
as to the formalism to be used for this. There is no lack of
competing competence grammars available, but also no reason to
expect that such grammars should be suited to a DOP approach,
designed as they were to characterize the nature of linguistic
competence rather than performance.

The next section sets out some of the properties that we might
require from such a ``performance grammar'' and offers a formalism
which attempts to
satisfy these requirements.

\vspace{-3pt}
\section{A Formalism for DOP}
\vspace{-5pt}

Given that we are attempting to construct a formalism that will do
justice to both the statistical and structural aspects of language,
the features that we would wish to maximize will include the
following:

\begin{enumerate}
\item
The formalism should be easy to use with probabilistic processing techniques,
ideally having a close correspondence to a simple probabilistic model
such as a Markov process.
\item
The formalism should be fine-grained, ie.\ responsive
to the behaviour of individual words (as n-gram models
are). This suggests a {\bf radically
  lexicalist} approach (cf.\ Karttunen, 1990) in which all rules are
encoded in the lexicon, there being no phrase structure rules which do
not introduce lexical items.
\item
It should be capable of capturing fully the linguistic intuitions of
language users. In other words, using the formalism one should be able to
characterize the structural regularities of language with at least the
sophistication of modern competence grammars.
\item
As it is to be used with real data, the formalism should be able to
characterize the wide range of syntactic structures found in actual
language use, including those normally excluded by competence grammars
as belonging to the ``periphery'' of the language or as being
``ungrammatical''. Ideally every interpretable utterance should
have one and only one analysis for any interpretation of it.
\end{enumerate}
Considering the first of these points, namely a close relation to a simple
probabilistic model, a good place to
start the search might be with a right-branching {\bf finite-state
grammar}. In this class of grammars every rule has the form
{\bf A} $\rightarrow$ {\bf a B} ({\bf A},{\bf B} $\in$
\{non-terminals\}, {\bf a} $\in$ \{terminals\})
and all trees have the simple structure :

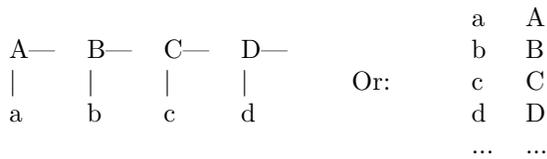
\begin{tabular}{llll}
A--- & B--- & C--- & D---  \\
$|$ & $|$ & $|$ & $|$ \\
a & b & c & d
\end{tabular}
\hspace{0.1in}
Or:
\hspace{0.1in}
\begin{tabular}{ll}
a & A \\
b & B \\
c & C \\
d & D \\
... &  ...
\end{tabular}
(with an
equivalent vertical alignment, henceforth to be used in this
paper, on the right)

In probabilistic terms, a finite-state grammar corresponds to a first-order
Markov process, where
 given a
sequence of states {\small S}$_i$, {\small S}$_j$,... drawn from a
finite set of possible states
\{{\small S}$_0$,..,{\small S}$_n$\} the probability of
a particular state occurring depends solely on the
identity of the previous state. In the finite-state grammar each word is
associated with a transition between two
categories, in the tree above {\bf `a'} with the transition
{\bf A} $\rightarrow$ {\bf B} and so on.
To calculate the probability that a string of words {\bf x$_1$, x$_2$,
  x$_3$,... x$_n$} has the parse represented by the string of
category-states {\bf S$_1$, S$_2$, S$_3$,...S$_n$}, we simply take the
product
of the probability of each transition: ie.\ \(\prod_{i=1}^{n}
P(x_{i}:  S_{i} \rightarrow S_{i+1})\).

In addition to satisfying our first criterion, a
finite-state grammar also fulfills the requirement that the
formalism be radically lexicalist, as by definition every rule
introduces a lexical item.

\vspace{-3pt}
\subsection{Accounting for Linguistic Structure}
\vspace{-3pt}

If a finite-state grammar is chosen however, the third
criterion, that of linguistic adequacy, seems to present an
insurmountable stumbling block.  How can such a simple formalism, in
which syntax is reduced to a string of category-states, hope to
capture even the basic hierarchical structure, the familiar ``tree
structure'', of linguistic expressions?

Indeed, if the non-terminals are viewed as atomic
categories then there is no way this can be done. If however, in line
with most current theories, categories are taken to be bundles of
features and crucially if one of these features has the value of a
{\bf stack of categories}, then this hierarchical structure can indeed
be represented.

Using the notation {\bf A [B]} to represent a state of basic category
{\bf A} carrying a category {\bf B} on its stack, the hierarchical
structure of the sentence:

(1) The man gave the dog a bone.

can be represented as:

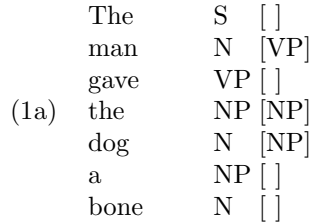
\begin{tabbing}
(1a))))\=boneeeeeee\=VPl\=\kill
\>The \>S  \>[ ] \\
\>man  \>N  \>[VP]  \\
\>gave   \>VP   \>[ ]  \\
(1a)\>the  \>NP  \>[NP]  \\
\>dog  \>N  \>[NP] \\
\>a  \>NP  \>[ ] \\
\>bone \>N   \>[ ]
\end{tabbing}

Intuitively, syntactic links between
non-adjacent words, impossible in a standard finite-state grammar, are
here established by passing categories along on the stack ``through''
the state of intervening words. That such a formalism can fully
capture basic linguistic
structures is confirmed by the proof in Aho (1968) that an {\bf
  indexed grammar} (ie.\ one where categories are supplemented with a
stack of unbounded length, as above), if
restricted to right linear trees (also as above), is equivalent to a {\bf
  context-free grammar}.

A perusal of the state transitions associated with individual words in
(1a) reveals an obvious relationship to the ``types'' of categorial
grammar. Using $\alpha$ to represent a list of categories (possibly
null), we arrive at the following transitions (with their
corresponding categorial types alongside).

The ditransitive verb `gave' is

\hspace{0.5cm} {\bf VP [$\alpha$]}\footnote{``VP'' is used here and
  henceforth as a shorthand for an S with a missing (ie.\ ``slashed'')
  subject.} $\rightarrow$ {\bf NP [NP,$\alpha$]} \hspace{0.6cm} (VP/NP)/NP

 Determiners in complement position are both:

\hspace{0.5cm}  {\bf NP [$\alpha$]} $\rightarrow$ {\bf N
  [$\alpha$]} \hspace{1.5cm} NP/N

 Determiner in subject position is
`type-raised'\footnote{The unidirectionality of the formalism results
  in an automatic type-raising of all categories appearing before
  their heads.} to:

\hspace{0.5cm} {\bf S [$\alpha$]} $\rightarrow$ {\bf N [VP,$\alpha$]}
\hspace{1.2cm} (S/VP)/N

The common nouns are all:

\hspace{0.5cm} {\bf N [$\alpha$]} $\rightarrow$ {\bf $\alpha$}
\hspace{2.5cm} N

In fact as no intermediate constituents are formed in the analysis, an
even closer parallel is to a dependency syntax where only rightward pointing
arrows are allowed, of which the formalism as presented above is a
notational variant. This lack of intermediate constituents has the
added benefit that no ``spurious ambiguities'' can arise.

Knowing now that the addition of a stack-valued feature suffices to
capture the basic hierarchical structure of language, additional
features can be used to deal with other syntactic relations. For
example, following the example of GPSG,  unbounded dependencies can be
captured using ``slashed'' categories. If we represent a ``slashed''
category {\bf X}  with the lower case {\bf x}, and use the
notation {\bf A(b)} for a category {\bf A} carrying a feature {\bf b},
then the topicalized sentence:

(2) This bone the man gave the puppy.

will have the analysis:

\begin{tabbing}
(2a))))\=puppyyyyyy\=VP(np)\=\kill
\>This \>S  \>[ ] \\
\>bone  \>N  \>[S(np)]  \\
\>the   \>S(np)   \>[ ]  \\
(2a)\>man  \>N  \>[VP(np)] \\
\>gave  \>VP(np)  \>[ ] \\
\>the  \>NP  \>[ ] \\
\>puppy \>N   \>[ ]
\end{tabbing}

Although there is no space in this paper to go into greater detail,
further constructions involving unbounded dependency and complement
control phenomena can be captured in similar ways.

\subsection{Coverage}

The criterion that remains to be satisfied is that of width of
coverage: can the formalism cope with the many ``peripheral''
structures found in real written and spoken texts?  As it stands the
formalism is weakly equivalent to a context-free
grammar and as such will have problems dealing with
phenomena like discontinuous constituents,
non-constituent coordination and gapping. Fortunately if extensions
are made to the formalism, necessarily taking it outside weak
equivalence to a
context-free grammar, natural and general analyses
present themselves for such constructions. Two of these will
now be sketched.

\subsection{Discontinuous Constituents}

Consider the pair of sentences  (3) and (4),
identical in interpretation, but the latter containing a discontinuous
noun phrase and the former not:

(3) I saw a dog which had no nose yesterday.

(4)  I saw a dog yesterday which had no nose.

which have the respective analyses:
\vspace{0.4cm}
\begin{tabbing}
(3a))))\=yesterdayyyy\=NP(t)\=[NP(t)]aa\= \kill
\>I \>S  \>[ ] \\
\>saw  \>VP  \>[ ]  \\
\>a   \>NP   \>[NP(t)]\>`t' =    \\
\>dog  \>N  \>[NP(t)]\>`time adjunct'  \\
(3a)\>which \>S(rel)   \>[NP(t)] \> `rel' =\\
\>had \>VP   \>[NP(t)] \> `relative' \\
\>no  \>NP  \>[NP(t)] \\
\>nose \>N   \>[NP(t)] \\
\>yesterday \>NP(t)   \>[ ]
\end{tabbing}

\begin{tabbing}
(4a)))))\=yesterdayyyy\=NP(t)\=[NP(t)]aa\= \kill
\>I \>S  \>[ ]  \\
\>saw \>VP \>[ ]  \\
\>a \>NP  \>[NP(t)] \\
\>dog \>N  \>[NP(t)]  \\
(4a)\>yesterday \>NP(t) \>[S(rel)]   \\
\>which \>S(rel)  \>[ ]  \\
\>had \>VP  \>[ ]  \\
\>no \>NP  \>[ ]   \\
\>nose \>N  \>[ ]
\end{tabbing}

The only transition in (4a) that differs from that of the corresponding word
in the
`core' variant (3a) is that of `dog' which has the respective
transitions:

\hspace{0.6cm} {\bf N [NP(t)]} $\rightarrow$ {\bf S(rel)
  [NP(t)]}\hspace{0.5cm} (in 3a)

\hspace{0.6cm} {\bf N [NP(t)]} $\rightarrow$ {\bf NP(t)
  [S(rel)]}\hspace{0.5cm} (in 4a)

Both nouns introduce a relative clause modifier {\bf S(rel)}, the
difference being that in the discontinuous variant a
category has been taken off the stack at
the same time as the modifier has been placed
on the stack. It has been assumed so far that we are using a right-linear
indexed grammar, but such a rule is expressly disallowed in
an indexed grammar and so allowing transitions of this kind ends the
formalism`s weak equivalence to the context-free grammars.

Of course, having allowed such crossed dependencies, there is nothing
in the formalism itself that will disallow a similar analysis for a
discontinuity unacceptable in English such as:

(5) I saw a yesterday dog.

This does not present a problem, however, as in DOP it is
information in the parsed
corpus which determines the structures that are possible. There is no
need to explicitly rule out (5), as the transition {\bf NP [$\alpha$]
$\rightarrow$ $\alpha$ [N]} will be vanishingly rare in any corpus of
even the most garbled speech, while the transition {\bf N [$\alpha$]}
$\rightarrow$ {\bf $\alpha$ [S(rel)]} is commonly met with in both
written and spoken English.

\vspace{-3pt}
\subsection{Non-Constituent Coordination}
\vspace{-3pt}

The analysis of standard coordination is shown in (6):

\begin{tabbing}
(6))))\=boneeeeeee\=VP(+)\=\kill
\>Fido \>S  \>[ ] \\
\>gnawed  \>VP  \>[ ]  \\
\>a   \>NP   \>[VP(+)]  \\
(6)\>bone  \>N  \>[VP(+)] \\
\>and \>VP(+)   \>[ ] \\
\>barked \>VP   \>[ ]
\end{tabbing}

Instead of a typical transition for `gnawed' of {\bf VP}
$\rightarrow$ {\bf NP}, we have a transition introducing a coordinated
VP: \hspace{0.2cm} {\bf VP} $\rightarrow$ {\bf NP [VP(+)]}

In general for any transition {\bf X } $\rightarrow$ {\bf Y },
where {\bf X} is a category and {\bf Y} a list of categories (possibly empty),
there will be a transition introducing coordination: \hspace{0.2cm}
{\bf X } $\rightarrow$ {\bf Y [X(+)]}

Non-constituent coordinations such as (7) present serious problems for
phrase-structure approaches:

(7) Fido had a bone yesterday and biscuit today.

However if we generalize the schema already obtained for
standard coordination by allowing {\bf X} to be not only a single
category, but a list of categories\footnote{There is in general no
  upper limit to the  length of this list, eg. ``I
  gave Fido a biscuit yesterday in the house and Rover a bone today in
  his kennel.''}, it is found to suffice for non-constituent
coordination as well.

\begin{tabbing}
(7a))))\=boneeeeeee\=NP(t)\=\kill
\>Fido \>S  \>[ ] \\
\>had  \>VP  \>[ ]  \\
\>a   \>NP   \>[NP(t)]  \\
(7a)\>bone  \>N  \>[NP(t)] \\
\>yesterday \>NP(t)   \>[N(+) [NP(t)]] \\
\>and \>N(+)   \>[NP(t)] \\
\>biscuit  \>N  \>[NP(t)] \\
\>today \>NP(t)   \>[ ]
\end{tabbing}

In this analysis instead of a regular transition for `bone' of:
\hspace{0.2cm} {\bf N [NP(t)]} $\rightarrow$ {\bf NP(t) [ ]}

there is instead a transition introducing coordination:
\hspace{0.2cm} {\bf N [NP(t)]} $\rightarrow$ {\bf NP(t) [N(+) [NP(t)]]}

Allowing categories on the stack to
themselves have non-empty stacks moves the formalism
one step further from being an indexed grammar. This is the final
incarnation of the formalism, being the {\bf
  State-Transition Grammar} of the title\footnote{Milward (1990)
  introduces a  formalism essentially identical to the
  one presented here, although viewed from a very different
  perspective. Milward (1994) shows how it handles a wide range
  of  non-constituent co-ordinations.}.

Similar schemas are being investigated to characterize {\bf gapping}
constructions.

\vspace{-3pt}
\subsection{Centre-Embedding}
\vspace{-3pt}

It should be noted that an indefinite amount of {\bf
  centre-embedding} can be described, but only at the
expense of unlimited growth in the length of states:

\begin{tabbing}
(8))))\=scratchedddd\=VP(np)\=\kill
\>The \>S  \>[ ] \\
\>fly  \>N  \>[VP] \\
\>the  \>S(np)  \>[VP] \\
\>dog \>N \>[VP(np),VP] \\
(8)\>the \>S(np)  \>[VP(np),VP] \\
\>cat \>N  \>[VP(np),VP(np),VP] \\
\>scratched \>VP(np)  \>[VP(np),VP] \\
\>swallowed \>VP(np)  \>[VP] \\
\>died \>VP  \>[ ]
\end{tabbing}

This contrasts with unlimited right-recursion where there is no
growth in state length:

\begin{tabbing}
(9))))\=scratchedddd\=S(rel)\=[VP]aaaa\=\kill
\>I \>S  \>[ ] \\
\>saw  \>VP  \>[ ] \\
\>the  \>NP  \>[ ] \\
\>cat  \>N  \>[ ] \\
(9)\>that  \>S(rel)  \>[ ] \\
\>scratched  \>VP  \>[ ] \\
\>the \>NP \>[ ] \\
\>dog \>N  \>[ ] \\
\>that \>S(rel)  \>[ ]  \\
\>... \>...  \> \\
\end{tabbing}
\vspace{-5pt}
As the model is to be trained from real data, transitions
involving long states as in (8) will have an ever smaller and eventually
effectively nil probability. Therefore, when tuned to any particular
language corpus the resulting grammar will be effectively
finite-state{\footnote{This may be compared to the claim in Krauwer \&
Des Tombes (1981) that finite-state automata offer a more satisfactory
characterization of language than context-free grammars.}.

\vspace{-5pt}
\section{Parsing}
\vspace{-5pt}

Assuming that we now have a corpus parsed  with the state-transition
grammar, how can this information be used to parse fresh text?

Firstly, for each word type in the corpus we can collect the
transitions with which it occurs and calculate its
probability distribution over
all possible transitions (an infinite number of which will be
zero).
To make this concrete, there are five tokens of the word `dog' in the
examples thus far, and so `dog' will have
the transition probability distribution:
\begin{tabbing}
{\bf N [VP(np),VP]} $\rightarrow$ {\bf S(np)  [VP(np),VP]]]}\=\kill
{\bf N [NP]} $\rightarrow$ {\bf NP [ ]}\>0.2 \\
{\bf N [NP(t)]} $\rightarrow$ {\bf S(rel) [NP(t)]}\>0.2 \\
{\bf N [NP(t)]} $\rightarrow$ {\bf NP(t) [S(rel)]}\>0.2 \\
{\bf N [VP(np),VP]} $\rightarrow$ {\bf S(np) [VP(np),VP]}\>0.2 \\
{\bf N [ ]} $\rightarrow$ {\bf S(rel) [ ]}\>0.2
\end{tabbing}

To find the most probable parse for a sentence, we simply find
the path from word to word which maximizes the product of the state
transitions (as we have a first order Markov process).

However this simple-minded approach, although easy
to implement, in other ways leaves much to be desired. The probability
distributions are far too ``gappy'' and even if a huge amount of data
were collected, the chances that they would provide the desired path
for a sentence of any reasonable length are slim. The process of
generalizing or smoothing the transition probabilities is therefore
seen to be indispensable.

\vspace{-3pt}
\subsection{Smoothing Probability Distributions}
\vspace{-3pt}
Although far from exhausting the possible methods for smoothing, the
following three are those used in the implementation described at the
end of the paper.

1. Factor out elements on the stack
which are merely carried over from state to state (which was done
earlier in looking at the correspondence of state transitions to
categorial types). The previous transitions for `dog'
then become:

\begin{tabbing}
aaaaaa\={\bf N [$\alpha$]} $\rightarrow$ {\bf S(rel)
[$\alpha$]]]]]]]]]]}\=\kill
\>{\bf N [$\alpha$]} $\rightarrow$ {\bf $\alpha$ [ ]}\>0.2 \\
\>{\bf N [$\alpha$]} $\rightarrow$ {\bf $\alpha$ [S(rel)]}\>0.2 \\
\>{\bf N [$\alpha$]} $\rightarrow$ {\bf S(np) [$\alpha$]}\>0.2 \\
\>{\bf N [$\alpha$]} $\rightarrow$ {\bf S(rel) [$\alpha$]}\>0.4
\end{tabbing}

2. Factor out other features which are merely
passed from state to state. For instance in the
example sentences, `the' has the generalized transitions:

\begin{tabbing}
aaaaa\=\kill
\>{\bf S [$\alpha$]} $\rightarrow$ {\bf N [VP,$\alpha$]} \\
\>{\bf S(np) [$\alpha$]} $\rightarrow$ {\bf N [VP(np),$\alpha$]}
\end{tabbing}

which can be further generalized to the single transition:

\begin{tabbing}
aa\={\bf S($\beta$) [$\alpha$]} $\rightarrow$ {\bf N
  [VP($\beta$),$\alpha$]]]]]}\=\kill
\>{\bf S($\beta$) [$\alpha$]} $\rightarrow$ {\bf N
  [VP($\beta$),$\alpha$]}\>$\beta$ = set of features
\end{tabbing}

3. Establish word paradigms, ie.\
classes of words which occur with similar
transitions. The probability distribution for individual words can
then be smoothed by suitably blending in the paradigmatic distribution. These
paradigms  will correspond to
a great extent to the word classes of rule-based grammars. The
advantage would be retained however that the system is still
fine-grained enough to reflect the idiosyncratic patterns of
individual words and could override this paradigmatic information if
sufficient data were available.

Words hitherto unknown to the system can be treated as being extreme
examples of words lacking sufficient transition data and they might then
be given a transition distribution blended from the open class word paradigms.

\vspace{-3pt}
\subsection{Problems Arising from Smoothing}
\vspace{-3pt}
Although essential for effective processing, the smoothing operations
may give rise to new problems. For example, factoring out items on the
stack, as in (1), removes from the model the disinclination for long
states inherent in the original corpus.  To recapture this discarded
aspect of the language, it would be sufficient to
introduce into the model a probabilistic penalty based on state
length. This penalty may easily be calculated
according to the lengths of states in the parsed corpus.

Not only would this allow the modelling of the restriction on
centre-embedding, but it would also allow many other ``processing''
phenomena to be accurately characterized. Taking as an example
``heavy-NP shift'',  suppose that the corpus contained two
distinct transitions for the word `threw', with the particle `out'
both before and after the object.

\begin{tabbing}
toldddd\=VP $\rightarrow$ NP, X(off)fffff\=\kill
threw  \>VP $\rightarrow$ NP, X(out)   \>prob: p1 \\
      \>VP $\rightarrow$ X(out), NP    \>prob: p2
\end{tabbing}

Even if p1 were considerably greater than p2, the cumulative negative effect of
the longer states in (10) would eventually lead to the model giving
the sentence with the shifted NP (11) a higher probability.

\begin{tabbing}
(10))))\=boneeeeeee\=VP(np)l\=\kill
\>I \>S  \>[ ] \\
\>threw  \>VP  \>[ ]  \\
\>the  \>NP  \>[X(out)]  \\
\>bacon  \>N  \>[X(out)]  \\
(10)\>that  \>S(rel)  \>[X(out)]  \\
\>Fido  \>S(np)  \>[X(out)]  \\
\>had  \>VP(np)  \>[X(out)]  \\
\>chewed  \>VP(np)  \>[X(out)]  \\
\>out   \>X(out)   \>[ ]
\end{tabbing}

\begin{tabbing}
(11))))\=boneeeeeee\=VP(np)l\=\kill
\>I \>S  \>[ ] \\
\>threw  \>VP  \>[ ]  \\
\>out   \>X(out)    \>[NP] \\
\>the  \>NP  \>[ ]  \\
(11)\>bacon  \>N  \>[ ]  \\
\>that  \>S(rel)  \>[ ]  \\
\>Fido  \>S(np)  \>[ ]  \\
\>had  \>VP(np)  \>[ ]  \\
\>chewed  \>VP(np)  \>[ ]
\end{tabbing}

\vspace{-3pt}
\subsection{Capturing Lexical Preferences}
\vspace{-3pt}

One strength of
n-gram models is that they can capture a certain amount of lexical
preference information.
For example, in a bigram model trained on sufficient data the
probability of the bigram `dog barked' could be expected to be
significantly higher than `cat barked', and this slice of
``world knowledge'' is something our model lacks.

It would not be difficult to make a small extension to the present
model to capture such information, namely by introducing an additional
feature containing the ''lexical value'' of the head of a
phrase. Abandoning the shorthand `VP' and representing a
subject explicitly as a ``slashed'' NP, a sentence with added lexical
head features would appear as:

\begin{tabbing}
(12))))\=boneeeeeee\=S(rel,np(dog))\=\kill
\>The \>S  \>[ ] \\
\>dog  \>N(dog)  \>[S(np(dog))]  \\
\>which   \>S(rel,np(dog))   \>[S(np(dog))] \\
(12)\>chased  \>S(np(dog)) \>[S(np(dog))]  \\
\>the \>NP(cat) \>[S(np(dog))]  \\
\>cat  \>N(cat)  \>[S(np(dog))]  \\
\>barked  \>S(np(dog))  \>[ ]
\end{tabbing}

In contrast to n-grams, where this sentence would cloud somewhat the
``world knowledge'', containing as it does the bigram `cat barked', the
added structure of our model allows the lexical preference to be
captured no matter how far the head noun is from the head verb. From
(12) the world knowledge of the system would be reinforced by the two
stereotypical transitions:

`chased'  \hspace{0.55cm} {\bf S(np(dog))}  $\rightarrow$ {\bf NP(cat)} \\
`barked' \hspace{0.5cm} {\bf S(np(dog))}  $\rightarrow$ {\bf [ ]}

\vspace{-3pt}
\section{Present Implementation}
\vspace{-3pt}

16,000+ running words from section N of the Brown corpus (texts N01-N08)
were hand-parsed using the state-transition grammar. The actual
formalism used was much fuller than the rather
schematic one given above, including many additional features such as case,
tense, person and number.
Transition probabilities were generalized in the ways
discussed in the previous section.

\vspace{-3pt}
\subsection{Results}
\vspace{-3pt}

100 sentences of less than 15 words were chosen randomly from other
texts in section N of the Brown corpus (N09-N14) and fed to the parser
without alteration.
Unknown words in the input, of which there were obviously many,  were
assigned to one of seven orthographic classes and given appropriate
transitions calculated from the corpus.

\begin{itemize}
\item
27 were parsed correctly, ie.\ exactly the same as the hand parse or
differing in only relatively insignificant ways which the model could
not hope to know\footnote{Such as the system postulating that
  ``Jess''  was a surname, as against the
  hand-parser's guess of a masculine first name.}.
\item
23 were parsed wrongly, ie.\ the analysis differed from the hand parse
in some non-trivial way.
\item
50 were not parsed at all, ie.\ one or more of the transitions
necessary to find a parse path was lacking, even after generalizing
the transitions.
\end{itemize}

\vspace{-3pt}
\subsection{Future Development}
\vspace{-3pt}
Although the results at present are extremely modest, it should be
borne in mind both that the amount of data the system has to work on
is very small and that the smoothing of transition probabilities is
still far from optimal. The present target is to achieve such a level
of performance that the corpus can be extended by
hand-correction of the parser output, rather than hand-parsing from
scratch.
Not only will this hopefully save a certain amount of
drudgery, it should also help to minimize errors and
maintain consistency.

A more distant goal is to ascertain whether the performance of the
model can improve after parsing new texts and processing the data
therein even without hand-correction of the parses, and if so what
the limits are to such ``self-improvement''.

\vspace{-3pt}
\section{References}
\vspace{-5pt}

{\sc Aho A.V.} 1968. Indexed Grammars. {\em
  Journal of the ACM}, 15: 647--671.

{\sc Bod, Rens} 1992. A Computational Model of
Language Performance: Data Oriented Parsing. {\em
  COLING-92}.

{\sc Karttunen L.} 1990.
  Radical Lexicalism. In Baltin \& Kroch (eds), {\em
    Alternative conceptions of phrase structure},
  Univ of Chicago Press, pp 43--65.

{\sc Krauwer, Steven \& Des Tombes, Louis} 1981. Transducers and
Grammars as Theories of Language. {\em Theoretical Linguistics}, 8,
173--202.

{\sc Milward, David} 1990. Coordination
in an Axiomatic Grammar. {\it COLING-90}.

{\sc Milward, David} 1994.
Non-constituent Coordination: Theory and Practice. {\it COLING-94}.

\end{document}